\documentclass{INTERSPEECH2023}

\interspeechcameraready 

\usepackage[utf8]{inputenc}
\usepackage[english]{babel}

\usepackage{xcolor,colortbl}
\usepackage{blindtext}
\usepackage{graphicx}
\usepackage{amsmath}
\usepackage[tableposition=top]{caption}
\usepackage{multirow}
\usepackage{subcaption}
\usepackage[normalem]{ulem}

\usepackage{tikz}
\usetikzlibrary{automata, arrows.meta, positioning}

\usepackage{pgfplots}

\makeatletter
\newcommand\footnoteref[1]{\protected@xdef\@thefnmark{\ref{#1}}\@footnotemark}
\makeatother

\renewcommand{\vec}{\mathbf}
\newcommand{\kaldi}{Kaldi}

\title{Hystoc: Obtaining word confidences for fusion of end-to-end ASR systems}
\name{Karel Beneš, Martin Kocour, Lukáš Burget}
\address{Brno University of Technology, Czechia}
\email{\{ibenes,ikocour,burget\}@fit.vutbr.cz}

\begin{document}

\maketitle
 
\begin{abstract}
\noindent End-to-end (e2e) systems have recently gained wide popularity in automatic speech recognition.
However, these systems do generally not provide well-calibrated word-level confidences.
In this paper, we propose Hystoc, a simple method for obtaining word-level confidences from hypothesis-level scores.
Hystoc is an iterative alignment procedure which turns hypotheses from an n-best output of the ASR system into a confusion network.
Eventually, word-level confidences are obtained as posterior probabilities in the individual bins of the confusion network.
We show that Hystoc provides confidences that correlate well with the accuracy of the ASR hypothesis.
Furthermore, we show that utilizing Hystoc in fusion of multiple e2e ASR systems increases the gains from the fusion by up to 1\,\% WER absolute on Spanish RTVE2020 dataset.
Finally, we experiment with using Hystoc for direct fusion of n-best outputs from multiple systems, but we only achieve minor gains when fusing very similar systems.

\end{abstract}

\section{Introduction}

The rising popularity of end-to-end (e2e) systems comes largely from their seemingly simple structure.
While the underlying neural architectures vary greatly~\cite{chan-las,li-jasper,gulati2020conformer}, the decoding process is generally very simple.
Oftentimes, it consists of iteratively finding the highest-scoring entry in a softmax layer (listen-attend-spell~\cite{chan-las}, LAS, and recurrent neural transducers \cite{graves-rnnt}, RNN-T), possibly even in parallel (connectionist temporal classification~\cite{graves-ctc}, CTC).
%Compared to a beam-search through a recognition network~\cite{mohri-wfst} in the hybrid DNN/HMM systems, these approaches are straight-forward.
These approaches are straightforward compared to a complicated beam-search decoding through a recognition network~\cite{mohri-wfst} in the hybrid DNN/HMM systems.

However, this simplicity comes at a cost:
The outputs of e2e systems rarely provide localized transcription variants or their scores.
It may be possible to extract some sort of localized probability information; however, the outputs of modern large classification networks are known to be miscalibrated in general~\cite{ovadia-trusting,guo-calibration}.
The nature of this localized information may vary in its very nature (e.g., the input-synchronous CTC assumes a specific relation between neural network outputs and final transcription, and it operates on a very different time-scale than label-synchronous LAS system) or in granularity (graphemic / sub-word / word level outputs), thus preventing direct use of this information for confidence estimation on a unified level, e.g.~words.

The lack of word-level confidences limits the utility of e2e system, as word-level confidences provide several benefits:
(1) They can be utilized by downstream applications~\cite{sperber-etal-2017-neural}, (2) they can be used in the scope of semi-supervised learning to filter out uncertain parts of the machine annotated data~\cite{vesely-semisup}, or (3) they can be used for finer fusion of multiple ASR systems~\cite{fiscus-rover}.

In this work, we propose to estimate word-level confidences and reap the benefits they provide by finding the common denominator of e2e systems:
They are trained to directly model the probability $P(\vec{y} | \vec{x})$ of
token sequence $\vec{y}$ being the correct transcription of the input audio $\vec{x}$.
Therefore, we assume that when a list of $N$ best transcription variants (hypotheses) is produced by an e2e system, a score $s_i \propto \log P(\vec{y}=\vec{h}_i | \vec{x})$ is available for each hypothesis $\vec{h}_i$.
Since this score $s_i$ is independent of the internal tokenization of $\vec{h}_i$ by the ASR system, any desired re-tokenization of the hypothesis can be taken to bridge the gap between different e2e systems and allowing for a unified approach.

In this paper, we propose Hystoc\footnote{URL annonymized.}\,--\,a simple alignment-based procedure for obtaining word-level confidences.
We show that the confidences obtained with Hystoc are surprisingly well calibrated and that they significantly improve gains from fusion of multiple e2e systems.
Finally, we extend Hystoc to directly fuse individual hypotheses from different systems.

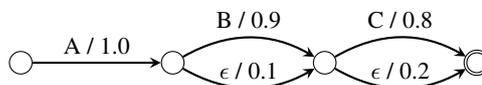
\begin{figure}[h!]
    \centering
    \begin{tikzpicture} [node distance = 2cm, on grid, auto]
     
        \node (q0) [state,inner sep=3pt,minimum size=0pt, initial text = {}] {};
        \node (q1) [state,inner sep=3pt,minimum size=0pt, right = of q0] {};
        \node (q2) [state,inner sep=3pt,minimum size=0pt, right = of q1] {};
        \node (q3) [state,inner sep=3pt,minimum size=0pt, right = of q2, accepting] {};
         
        \path [-stealth, thick]
            (q0) edge node {A / 1.0} (q1)
            ;

        \path [-stealth, thick]
            (q1) edge[bend left] node {B / 0.9} (q2)
            ;

        \path [-stealth, thick]
            (q1) edge[bend right] node {$\epsilon$ / 0.1} (q2)
            ;

        \path [-stealth, thick]
            (q2) edge[bend left] node {C / 0.8} (q3)
            ;

        \path [-stealth, thick]
            (q2) edge[bend right] node {$\epsilon$ / 0.2} (q3)
        ;
    \end{tikzpicture}
    \caption{
        A confusion network representing strings ABC, AB, and AC with probabilities 0.7, 0.2, and 0.1 respectively.
        Note that while each $\epsilon$-transition corresponds to a single hypothesis omitting the corresponding letter, all of the letter transitions actually aggregate multiple hypotheses.
    }\label{fig:my_label}
\end{figure}

\section{Distilling word-level confidences}

In this section, we describe how Hystoc\footnote{\textbf{HY}pothesis \textbf{S}cores to \textbf{TO}ken \textbf{C}onfidences} restores word-level confidences from hypothesis level scores.
Prior to the actual computation, observe that all hypotheses $h_i$ can be re-tokenized into a desired target form, e.g.\ into whitespace-delimeted words for European languages.

Hystoc is an iterative alignment procedure.
Specifically, it aligns individual hypotheses $h_i$ into a confusion network~\cite{bangalore_computing_2001} (see Fig.~\ref{fig:my_label}) in the order of decreasing hypothesis scores $s_i$.
When aligning $h_i$ into the confusion network, Hystoc finds the highest-scoring path $h^{*}$ through the confusion network and finds a Levenshtein alignment between $h_i$ and $h^{*}$.
Thus, the bins in the confusion network are determined into which the individual tokens from $h_i$ shall be incorporated, possibly introducing new bins in case the alignment contains insertions.
The score of each transition corresponding to a token from $h_i$ is then increased\footnote{In log-domain, i.e., by log-add-exp.} by $s_i / T$.
Here, $s_i$ is the score of the hypothesis $h_i$ and $T$ is \emph{temperature} --- a hyperparameter of Hystoc.

Finally, we normalize the scores in each bin by applying softmax.
The confidence of any word is then its probability within the corresponding bin.

If the outputs of the ASR system are to be rescored by any sort of external language model (LM), it is sufficient to incorporate the LM probabilities into the scores $s_i$ of the hypotheses prior to the application of Hystoc.

\subsection{Related work}
Rover --- a standard tool for fusing outputs from multiple ASR systems~\cite{fiscus-rover} --- operates similar to Hystoc.
However, Rover already expects word-level confidences in the input.
When an insertion/deletion mismatch is detected, Rover introduces an $\epsilon$-transition with a fixed confidence $C_\epsilon$.
Hystoc only assumes hypothesis level scores at the input and derives the scores of possible $\epsilon$-transitions individually.

There is a multitude of methods for estimating word-level confidences for e2e ASR systems~\cite{oneata-confidences}, but these typically include evaluating multiple models or training whole new models for confidence estimation.
Hystoc operates on the outputs of a single system.

Already in 1996, Wessel at al.\ have observed that word-level confidences derived from $N$-best list are well correlated to the error rates of hybrid GMM-HMM systems~\cite{wessel-confidence}.
Hystoc updates this observation for neural e2e systems and utilizes this information for effective system fusion.

Kišš has recently shown that hypothesis level scores are well correlated with segment-level error rate for character-level CTC systems in OCR and serve as an efficient filtration criterion for semi-sepervised training~\cite{kiss-at-st}.
Hystoc derives token-level confidences for a tokenization independent of the underlying neural model.

\subsection{Hystoc fusion}\label{sec:nbest-fusion}
The primary application aim of this work is to enable fusion of generic end-to-end systems, via the standard Rover tool.
Nevertheless, the core procedure of Hystoc is unaware of the source of the individual hypotheses that are aligned together.
Therefore, we can apply Hystoc confidence estimation to accumulation of hypotheses across systems, under the very mild condition that they are run on the same VAD segmentation.
This way, a system fusion is performed.
We hypothesise that this comes with the added benefit of also considering all the other hypotheses as opposed to only considering a 1-best output from each system.

We propose three strategies of incremental alignment:

\noindent\textbf{Direct},\quad where we take every hypothesis $h_i^s$ along with its associated score $s_i^s$ as-is and treat them as coming from a single system.
This assumes that the scores coming from all systems have comparable dynamic range.

\noindent\textbf{Normalized},\quad where we first normalize the scores for each system $s'$ separately so that $\sum_i\exp(s_i^{s'}) = 1$.
Then, we again drop the notion of individual systems and align the hypotheses into the confusion network in the order given by their scores.
This way, we roughly balance the scale of scores across systems, but allow more confident systems to impact the confusion network more by having their top hypotheses aligned into it first.

\noindent\textbf{Normalized round-robin}\quad addresses this aspect and first introduces the top-1 hypothesis from each system, then the second best from each system and so on.

\section{Experiments}
\begin{table}[]
    \centering
    \caption{%
        Performance of the 1-best output of each individual system, measured by \% WER.
    }\label{tab:single-system}
    \begin{tabular}{llrr}
        \toprule
         System     &   & w/o LM    & w/ LM \\
         \midrule
         ConfA      & A             & 17.3      & 16.7 \\
         ConfB      & B             & 17.7      & 17.1 \\
         RNN-T      & R             & 21.7      & 21.2 \\
         \kaldi     & K             & 22.1      & 20.0 \\
         \bottomrule
    \end{tabular}
\end{table}

We study the behaviour of Hystoc on the Albayzin 2022 challenge.
This speech recognition challenge comes with a collection of three audio databases:
RTVE2018 and RTVE2020, consisting of recordings from various Spanish TV shows broadcasted between 2015 and 2019, and RTVE2020, comprised of audio material from historical recordings, popular broadcasted TV shows and fictional shows~\cite{RTVE2022}.
We followed the original data splits and used the RTVE2022's dev partition with 2.5 hours of audio for development and RTVE2020's test partition with 39 hours for cross-validation of acoustic models. 
The remaining 738 hours were used for training.

Most of the provided training data are not human-revised, so we followed a data cleaning procedure from a previous work~\cite[Section 2.1]{kocour21_iberspeech}, resulting in 512 hours of training recordings and 41 minutes of well-annotated dev data. 
The cross-validation set remained untouched in our experiments.

% \todo{The winning team of the challenge delivered a fusion of several systems~\cite{kocour-albayzin}.
% Thus, we see it fit to replicate their setup as closely as possible for this study.
% In the following subsection, we give a high-level description of the system.
% For more technical aspects such as data augmentation and training details, the system description of BCN2BRNO team can be referred.}

\subsection{Single systems}

Since the main evaluation method of this work is system fusion, we work with multiple ASR systems operating on a shared segmentation of the audio data.
This segmentation was obtained using a voice activity detection in the form of a simple feed-forward neural network processing 31 consecutive frames represented by 15 Mel-filter bank features alongside with 3 Kaldi pitch features~\cite{gharemani:ICASSP:2014:kaldi_pitch}.

In total, we consider four different ASR models:

\noindent\textbf{ConfA}\quad
An XLS-R Conformer~\cite{guo_conformer_espnet2020} consisting of 12 encoder and 6 decoder layers.
The input features are extracted from a pre-trained XLS-R wav2vec2.0~\cite{babu22_interspeech}.
We use the 0.3\,billion parameters version of XLS-R trained on 128 languages.
As outputs, we use 1500 byte pair encoding (BPE) units.
The Conformer model was trained from scratch on the training data, while the XLS-R feature extractor remained frozen.

\noindent\textbf{ConfB}\quad
An XLS-R Conformer of the same architecture as ConfA, but trained additionally on 400 hours of Spanish Common Voice~\cite{ardila-common-voice}.

\noindent\textbf{RNN-T}\quad
A recurrent transducer~\cite{graves_rnn-t_2012} based on CRDNN~\cite{speechbrain} encoder and GRU~\cite{revanelli_gru_2018} prediction network.
This model operates on 80-dimensional filter banks and predicts a vocabulary of 1000 BPE tokens.

\noindent\textbf{\kaldi}\quad
A hybrid DNN/HMM system implemented in Kaldi.
We utilize the LF-MMI objective function~\cite{Povey:IS:2019:LF-MMI} and we have our acoustic model predict biphones.
The architecture is a 19-layer TDNN-f~\cite{povey:IS:2018:TDNN-F} preceded by a 6-layer CNN front-end.
This model operates on 40-dimensional Mel-filter bank features and is additionally presented with x-vectors~\cite{karafiat-xvector} as means of speaker adaptation.
A 3-gram language model is used for decoding, its optimal weight is 0.7.

Furthermore, we train an external language model (LM) based on the LSTM architecture~\cite{sundermeyer12_interspeech}, with two layers of 1500 units each.
The LM operates on a 20k BPE vocabulary.
We pretrained it on Spanish News Crawl monolingual texts prepared for WMT'13 challenge\footnote{\url{https://www.statmt.org/wmt13/translation-task.html}} and fine-tuned it to transcripts of the training data.
The LM is utilized by rescoring the 100-best outputs from each system.
When rescoring the Kaldi system, we perform a log-linear interpolation between the scores from the LSTM-LM and the original 3-gram LM.
The optimal ratio is 0.7 of LSTM-LM to 0.3 of the 3-gram model. The interpolation is rather insensitive to the exact value and LSTM-LM weights in the range $\langle{}0.6, 0.8\rangle$ give virtually the same performance.
When rescoring the e2e systems, we add the LM probability to the ASR score in the log-domain and introduce a \emph{token insertion bonus}, i.e.\ an increase in log-probability of a hypothesis proportional to its length as measured in number of LM tokens.
Here, the optimal values are in range $\langle{}0.2, 0.3\rangle$ for LM weight and $\langle{}5.5, 6.5\rangle$ for token insertion bonus.

The performance of our ASR systems, both with and without LM rescoring is summarized in Table~\ref{tab:single-system}. 
Note that rescoring of $N$-best lists improved WER of each system by 0.7\,\% absolute on average.
The XLS-R Conformer model (ConfA) achieved 16.7\,\% and 17.3\,\% WER with and without LM rescoring on RTVE2020 test set.
These results are comparable with the results obtained by the winners of Albayzin 2022 Speech to Text Challenge~\cite{kocour-albayzin}.

\subsection{Quality of confidences}

\begin{figure}
    \centering
    \includegraphics[width=\linewidth]{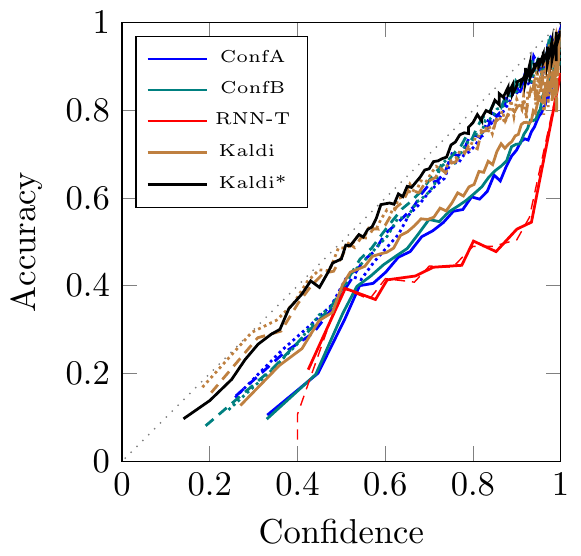}
    \caption{
        Confidence as a predictor of accuracy.
        Dashed line corresponds to using temperature $T=3.0$, dotted to $T=10.0$.
        \kaldi* is the set of confidences coming directly from the original Kaldi lattices, i.e.\ without their reduction to 100-best.
        Each coordinate is computed from 2500 words.
    }
    \label{fig:conf-acc}
\end{figure}

\begin{figure}
    \centering
    \includegraphics[width=\linewidth]{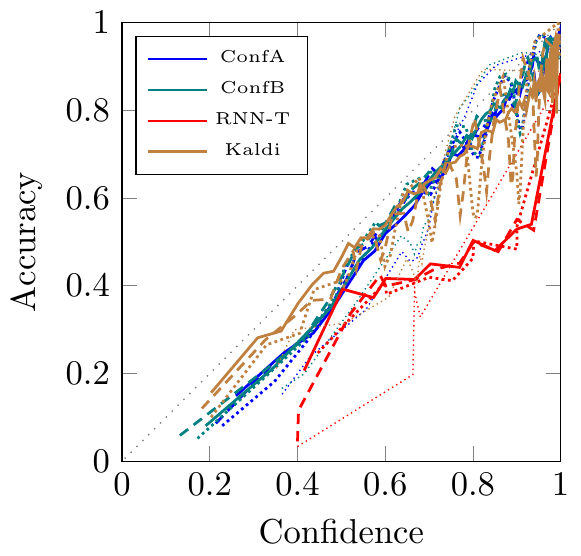}
    \caption{
        Confidence quality as a function of number of top scoring hypotheses considered by Hystoc.
        Thinly dotted, dotted, dashed and full lines correspond respectively to 3-, 10-, 17- and 100-best outputs from each system.
    }
    \label{fig:conf-nb-best}
\end{figure}

With the ASR systems in place, we explore the quality of the confidence as a predictor of the word accuracy.
To this end, we process the dev data with an ASR system and for each token, we record its confidence and whether it aligns correctly to the ground truth.
Then, we sort the tokens by confidence and group them into batches of 2500 tokens.
For each of these batches, we plot its median confidence and the proportion of correctly aligned tokens in the batch.
Note that this analysis correctly takes into account substitution and insertion errors, but it does not capture deletions.

The results of this analysis are shown\footnote{For systems w/o LM rescoring. The trends are very similar for systems with the rescoring.} in Figure~\ref{fig:conf-acc}.
With the exception of the RNN-T model, all confidences are in a close linear relationship with the word accuracies in the corresponding cohorts.
With an elevated temperature $T=3.0$, the dependency is even closer to the optimum.
Increasing the temperature further to $T=10.0$ brings no additional gain.

Increasing the temperature means giving more credibility to hypotheses which were originally scored as less likely, so we investigated the WERs of these confusion networks.
The change in WER caused by increasing $T$ from $1.0$ to $3.0$ was smaller than 0.1\,\% absolute and it was even negative in majority of cases.
The one outlier was the ConfA system rescored with the external LM, where the WER increased by 0.2\,\% absolute.
Overall, we conclude that it is safe to increase the temperature if one desires more precise estimates of word accuracy.

Furthermore, we check how is the relationship between confidence and word accuracy influenced by the number of $N$-best outputs.
Figure~\ref{fig:conf-nb-best} shows that while applying Hystoc to 3-best is insufficient, the confidences are already reasonably reliable with as little as 10-best outputs from an ASR.
Again, with the exception of the RNN-T system.

\subsection{Advantage of confidences in Rover fusion}
\begin{table}[]
    \centering
    \caption{%
        Performance of Rover fusion as a function of temperature $T$ used to obtain the per-system confidences; results reported as \% WER.
        Note that with $T=0$, only the best path is considered from each system and the resulting confidences are then all equal to 1.0.
    }\label{tab:rover-fusion}
    \begin{tabular}{llrrr}
        \toprule
         LM     & Systems        & \multicolumn{3}{c}{$T$} \\
                &               & 0.0        & 1.0  & 3.0 \\
         \midrule
         no    & A              & 17.3     & 17.3   & 17.3 \\
               & A + B          & 17.4     & 16.8   & 16.8 \\
               & A + B + K      & 16.7     & 16.2   & 16.2 \\
               & A + B + R      & 16.9     & 16.5   & 16.5 \\
               & A + B + K + R  & 16.1     & 15.9   & 16.0 \\
         \midrule
         yes   & A              & 16.7     & 16.7   & 16.9 \\ 
               & A + B          & 17.2     & 16.5   & 16.5 \\
               & A + B + K      & 16.4     & 15.4   & 15.7 \\
               & A + B + R      & 16.4     & 15.8   & 16.0 \\
               & A + B + K + R  & 15.8     & 15.3   & 15.6 \\
         \bottomrule
    \end{tabular}
\end{table}

Observing that the obtained per-word confidences are reasonable predictors of word accuracy, we proceed to leverage them in fusion of multiple systems.
To this end, we run the standard Rover fusion on transcriptions with Hystoc confidences.
As the natural baseline, we take the single-best hypothesis from each system and assign confidence 1.0 to each word in it.
Formally, we achieve this effect by setting the temperature to $T = 0$.
Motivated by the effect of temperature observed in the previous section, we introduce one more setup into this experiment, where we increase the temperature of each system correspondingly to 3.0.
% We sweep over Rover the hyperparameters, the voting\,--\,confidence interpolation coefficient $\alpha$ and $\epsilon$-confidence $C_\epsilon$ in a range of $\langle{}0, 1\rangle$.
We tune the Rover fusion hyperparameters, i.e. confidence interpolation coefficient $\alpha$ and $\epsilon$-confidence $C_\epsilon$, in a range of $\langle{}0, 1\rangle$.

The results of this experiment are summarized in Table~\ref{tab:rover-fusion}.
Providing the Hystoc confidences is beneficial in all cases, though the scale of the improvement varies.
In general, the improvement is roughly on the scale of having one more system included in the fusion, i.e., the fusion of $N$ systems with the Hystoc confidences is about as accurate as fusion of $N+1$ systems without confidences.
This is in line with the observations made in the Rover paper~\cite{fiscus-rover} for hybrid systems.
Interestingly, the improvement from Hystoc confidences is better pronounced for systems with external LM rescoring.

While the optimal values of $\alpha$ were consistently around $0.55$ for $T=1.0$, they increased to about $0.75$ for $T=3.0$.
This suggests that the fusion mechanism generally searches for a smoothed but informative confidence distributions.

\subsection{Replacing Rover fusion with Hystoc}
\begin{table}[]
    \centering
    \caption{%
        Performance of Hystoc fusion with different schemes of accumulating individual hypotheses, reported as \% WER.
        ``Norm-RR'' stands for normalized round-robin.        
    }\label{tab:nbest-fusion}
    \begin{tabular}{llrrr}
        \toprule
         LM     & Systems        & \multicolumn{3}{c}{Scheme} \\
                &               & Direct   & Normalized & Norm-RR \\
         \midrule
         no    & A              & 17.3     & 17.3       & 17.3 \\
               & A + B          & 17.0     & 17.0       & 17.2 \\
               & A + B + K      & 17.0     & 17.1       & 17.3 \\
               & A + B + R      & 21.5     & 16.9       & 16.9 \\
               & A + B + K + R  & 21.6     & 19.6       & 19.5 \\
         \midrule
         yes   & A              & 16.7     & 16.7       & 16.9 \\ 
               & A + B          & 16.8     & 16.3       & 16.4 \\
               & A + B + K      & 16.7     & 16.5       & 16.5 \\
               & A + B + R      & 18.7     & 16.1       & 16.1 \\
               & A + B + K + R  & 18.9     & 19.2       & 19.2 \\
         \bottomrule
    \end{tabular}
\end{table}

As described in Section~\ref{sec:nbest-fusion}, we continue to explore the possibility of fusing ASR outputs directly on level of aligning individual hypotheses.
We fuse the same sets of models as in the previous experiment, making the results directly comparable.
As the elevated temperature does not consistently improve the results, we use $T=1.0$.
With Hystoc fusion, there are no more hyperparameters to tune.

The results are shown in Table~\ref{tab:nbest-fusion} and overall, they are considerably worse than in the Rover fusion.
As long as we do not include the RNN-T, the fusion is at least slightly better than using the single best system.
When RNN-T is included, the WERs deteriorate dramatically, esp.~when LM rescoring is not applied.
While we do not have a solid explanation why this happens for RNN-T in particular, it is in line with the observation that the confidences derived from RNN-T are not very well calibrated.
This suggests that the confidence--accuracy analysis is a reasonable predictor of the usefulness of a system in the fusion. 

One notable exception is the fusion of the two conformer systems with LM, where the Hystoc fusion yields slightly better results.
While not significant on its own, this result does provide some hope for further improvements of fusion of similar systems.

\section{Conclusion}
In this paper, we proposed Hystoc, a straight-forward technique for computing word-level confidences from $N$-best lists produced by any end-to-end system.
Hystoc is fully independent of both the tokenization and the exact objective formulation used by the underlying ASR model, making it a very generic tool.

We have shown that the confidences estimated by Hystoc are on-par with confidences obtained from lattices from a WFST-based decoder, and with as few as 10-best, the confidence estimation is reasonably stable.
Practically, we have shown that Hystoc confidences bring a significant gain into the fusion of ASR systems, yielding 0.2 to 0.6\,\% absolute improvement over a confidence-free voting scheme.
Finally, we experimented with a direct Hystoc fusion of $N$-best lists from different systems.
Here, the results are negative, we only obtained minor gains when fusing a specific combination of models, while for most other setups, the performance deteriorated.

We see two lines of further work on Hystoc:
First, when aligning the individual hypotheses, Hystoc drops the notion of sequences immediately and treats tokens individually, allowing ``hypothesis hopping'' at any position.
It could be beneficial to devise a more fine-grained approach where it would only be possible to switch between hypotheses in positions of high disagreement.
Second, the iterative nature of hypothesis alignment in Hystoc introduces hard decisions early in the process of constructing the confusion network, possibly limiting their quality.
We expect that doing a full multi-sequence alignment might mitigate this issue --- however its direct computation is prohibitively expensive.

\section{Acknowledgements}

The authors would like to thank the corresponding grant agencies for the projects that provided computing power for and covered funding of this work.

\bibliographystyle{IEEEtran}
\bibliography{mybib}

\end{document}